\documentclass[letterpaper,twocolumn,fleqn]{article}

\usepackage{ist,graphicx,subfigure,amsmath,amssymb}
\graphicspath{{figures/}}

\title{Interactive Hand Pose Estimation: Boosting accuracy in localizing extended finger joints}

\author{Cairong Zhang, Guijin Wang, Hengkai Guo, Xinghao Chen, Fei Qiao, Huazhong Yang; \\Department of Electronic Engineering, Tsinghua University; Beijing, China}

\date{} % date has an empty field.

\begin{document}

\maketitle

\thispagestyle{empty} % prevents the first page to be numbered

%%%%%%%%%%%%%%%%%%%%%%%%%%%%%%%%%%
% Abstract
%%%%%%%%%%%%%%%%%%%%%%%%%%%%%%%%%%

\begin{abstract}
Accurate 3D hand pose estimation plays an important role in Human Machine Interaction (HMI). In the reality of HMI, joints in fingers stretching out, especially corresponding fingertips, are much more important than other joints. We propose a novel method to refine stretching-out finger joint locations after obtaining rough hand pose estimation. It first detects which fingers are stretching out, then neighbor pixels of certain joint vote for its new location based on random forests.

The algorithm is tested on two public datasets : MSRA15 and ICVL. After the refinement stage of stretching-out fingers, errors of predicted HMI finger joint locations are significantly reduced. Mean error of all fingertips reduces around 5mm (relatively more than 20\%). Stretching-out fingertip locations are even more precise, which in MSRA15 reduces 10.51mm (relatively 41.4\%).
\end{abstract}

\section{Keywords}
Random Forest, Hand Pose Estimation, Human Machine Interaction, Fingertip Detection

\section{1. Introduction}
\label{secion_introduction}

Hand pose estimation plays an important role in Human Machine Interaction (HMI), such as virtual reality (VR), augmented reality (AR) and remote control. Estimating hand poses from individual depth images has drawn lots of attention from researchers \cite{xu2013efficient, tang2014latent, sun2015cascaded, supancic2015depth, wan2016hand, ge2016robust, guo2017region}, thanks to the availability of depth
cameras \cite{zhang2012microsoft, wang2013depth, shi2015high, keselman2017intel}, such as Microsoft Kinect, Intel Realsense Camera etc.

Recently convolutional neural networks \cite{tompson2014real, oberweger2015hands, oberweger2015training, ge2016robust, sinha2016deephand, guo2017region} and random forests \cite{tang2014latent, sun2015cascaded, tang2015opening, li20153d, supancic2015depth, wan2016hand} are two mainstreams in hand pose estimation. The performances of these two methods are roughly comparable. But random forests can achieve faster prediction without any GPUs, so it is beneficial to implement online real-time application with only CPUs based on random forests.

Joints in palm such as the wrist and five finger roots have lower degree of freedom (DOF) than those in fingers. Considering kinematic constraints of hand joints, \cite{sun2015cascaded} proposed a hierarchical regression method, which first regresses locations of joints with relatively low DOF and then those with higher DOF. He also introduced a cascaded framework to refine joint locations iteratively.

However, due to extremely high fingertip flexibility, similar part confusion, large range of movement and poor depth quality around fingertips, the errors of fingertip locations are improperly high, almost always the highest in all joints. In the reality of HMI, crucial actions such as clicking and zooming depend heavily on stretching-out fingers, especially extended fingertips. Therefore, more accurate estimation of stretching-out finger joints, especially fingertips, is in urgent need.

In this work, we present a novel scheme named Interactive Hand Pose Estimation (IHPE) to address the conflicts between HMI demands and large errors of fingertip locations in hand pose estimation. We first use cascaded hierarchical regression in \cite{sun2015cascaded} to get rough locations of hand joints. This module is implemented by reproducing corresponding algorithms in \cite{sun2015cascaded} as our baseline, including two kind of stages: palm stages and finger stages. Inspired by fingertip detection and corresponding root localization proposed in \cite{chen2016static}, we then introduce a refinement stage to re-estimate joint locations of stretching-out fingers. In the refinement stage, we first detect where the stretching-out fingers are, based on \emph{Key Joints Localization} algorithms in our previous work \cite{chen2016static}. Then we incorporate the information provided by prediction of our baseline, to distinguish which fingers these detected extended fingers are. Finally, for each joint to be re-estimated, its neighbor foreground pixels (the pixels inside the hand region after segmentation are defined as foreground pixels) vote for its location using random forests. The average result of all votes won by the joint is its new location after refinement. As shown in our experiments, after the refinement stage of stretching-out finger joints, the errors of stretching-out finger joint locations are significantly reduced, especially the fingertips.

Our main contribution is the refinement method proposed to improve estimating locations of extended finger joints, particularly fingertips, after getting some poor results while estimating all joints. The accuracy gap between estimating locations of extended fingertips and other joints can be remarkably reduced after our refinement stage. It benefits the users by providing them more satisfying experience in HMI applications, such as VR and AR.

\section{2. Cascaded Hierarchical Regression}\label{section_baseline}
In this section, we give a brief introduction of our baseline: Cascaded Hierarchical Regression.

\begin{table}[htb]
  \caption{Table.1. Our baseline training procedure}
  \label{baseline_train}
  \begin{center}
  \begin{tabular}{p{0.02\columnwidth}p{0.96\columnwidth}}
    \hline
    \multicolumn{2}{l}{\emph{Algorithm 1}\hspace{2mm}Training algorithm for our baseline} \\
    \hline
    1. & \emph{input:} depth image $I_i$, ground truth pose ${\theta}_i$, and initial palm pose ${\theta}^{p, 0}_i$ for all training samples $i$ \\
    2. & \emph{for} $t = 1$ \emph{to} $T_1$ \emph{do} \hfill\emph{[palm stages]} \\
    3. & {\quad}${\delta}{\theta}^p_i = {\theta}^p_i - {\theta}^{p, t-1}_i$ \\
    4. & {\quad}learn random forest $R^{p, t}$ to approximate ${\delta}{\theta}^p_i$ \\
    5. & {\quad}${\theta}^{p, t}_i = {\theta}^{p, t-1}_i + R^{p, t}(I_i, {\theta}^{p, t-1}_i)$, update palm joint locations \\
    \\
    6. & initialize finger poses ${\theta}^{f, 0}_i (f = 1, 2, 3, 4, 5)$, so that five fingers are in the direction of a vector pointing from wrist (or palm center) to middle finger root \\
    7. & \emph{for} $t = 1$ \emph{to} $T_2$ \emph{do} \hfill\emph{[finger stages]} \\
    8. & {\quad}\emph{for} $f = 1$ \emph{to} $5$ \emph{do} \\
    9. & {\qquad}${\delta}{\theta}^f_i = {\theta}^f_i - {\theta}^{f, t-1}_i$ \\
    10. & {\qquad}learn random forest $R^{f, t}$ to approximate ${\delta}{\theta}^f_i$ \\
    11. & {\qquad}${\theta}^{f, t}_i = {\theta}^{f, t-1}_i + R^{f, t}(I_i, {\theta}^{f, t-1}_i)$, update finger joint locations \\
    12. & {\quad}${\theta}^t_i = {\theta}^{p, T_1}_i{\bigcup}\{{\theta}^{f, t}\}_{f=1,2,3,4,5}$ update all joints \\
    \\
    13. & \emph{output:} $\{R^{p, t}\}_{t=1,2,\cdots,T_1}, \{R^{f, t}\}_{t=1,2,\cdots,T_2; f=1,2,\cdots,5}$ \\
    \hline
  \end{tabular}
  \end{center}
\end{table}

As clarified in \cite{sun2015cascaded}, joints in different parts of the hand have large variance of flexibilities, so regressing all joints together may cause unnecessarily high complexity and degrade model's performance. We divide the whole regression task into two subtasks, palm stages and finger stages, as Sun. do in \cite{sun2015cascaded}. In palm stages, we regress locations of only those joints in palm such as the wrist, finger roots and the palm center. Then we regress joints in five fingers respectively with locations of joints in palm fixed. Besides, we use cascaded architecture in both palm and finger stages like \cite{sun2015cascaded} does. Specifically in this paper, we assign three stages to both palm regression and finger regression. Each stage produces coordinate differences between ground truth and predictions of last stage (Eq.\eqref{eq_cascaded}). Details of the training algorithm is shown in Table.\ref{baseline_train}. While testing, the output models of the training algorithm are used to perform regression in a similar procedure as training.
\begin{equation}\label{eq_cascaded}
  {\delta}{\theta}^t = {\theta}_{gt} - {\theta}^{t-1}
\end{equation}
Note that we don't use \emph{3D Pose Normalization} or \emph{3D Pose Indexed Features} as in \cite{sun2015cascaded}. We use traditional pixel difference features instead (Eq.\eqref{eq_feature}).

\section{3. Stretching-Out Finger Joint Refinement}\label{section_refinement}
We introduce our refinement of stretching-out finger joint locations in this section. It is divided into two stages. First we detect which fingers are stretching out. Then for each joint in these fingers, its neighbour pixels vote for its location through random forests.

\subsection{3.1 Stretching-out fingers detection}\label{section_stretching}
\begin{figure}[htb]
    \centering
    \subfigure[]{
      \label{fig_gesture_a}
      \includegraphics[width=0.20\columnwidth]{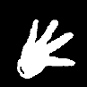}}
      \hspace{0.02\columnwidth}
    \subfigure[]{
      \label{fig_gesture_b}
      \includegraphics[width=0.20\columnwidth]{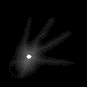}}
    \subfigure[]{
      \label{fig_gesture_c}
      \includegraphics[width=0.20\columnwidth]{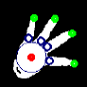}}
      \hspace{0.02\columnwidth}
    \subfigure[]{
      \label{fig_gesture_d}
      \includegraphics[width=0.20\columnwidth]{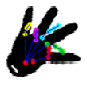}}
    \caption{Stretching-out fingers detection. (a)Hand mask. (b)Distance transform, the brightest point is palm center. (c)Detected finger tips and roots. (d)Prediction of the baseline.}
    \label{fig_gesture}
\end{figure}

We detect stretching-out fingers based on hand mask images (Fig.\ref{fig_gesture_a}), using the method proposed in our previous work \cite{chen2016static}. As clarified in \cite{chen2016static}, palm center can be defined as the point with maximal distance to the nearest palm boundary point (Fig.\ref{fig_gesture_b}). After the palm center is localized, it scans the palm boundary and get distances between boundary points and palm center. Those points with local maximal distance are stretching-out fingertips, if their distance exceeds a global threshold and the curvature is large enough. Then corresponding finger roots can be localized based on locations and fingertips and the palm center (Fig.\ref{fig_gesture_c}).

Then we obtain rough locations of other joints in these detected stretching-out fingers through interpolation according to finger root and tip locations. To distinguish which these extended fingers are, index, ring or others, we calculate the distances between each detected finger and five fingers predicted from our baseline (Fig.\ref{fig_gesture_d}) respectively. Each detected stretching-out finger is distinguished as the one with the minimum distance to it (Eq.\eqref{eq_fingers}).
\begin{equation}\label{eq_fingers}
  F = {\arg\min}_{f = 1,2,3,4,5}\sum_{j\in\theta^f}(\|\theta^f_{mask, j} - \theta^f_{baseline, j}\|^2_2)
\end{equation}
where $\theta^f_{mask, j}$ is location of the $j^{th}$ joint in finger $f$ from stretching-out fingers detection, $\theta^f_{baseline, j}$ is the location of the $j^{th}$ joint in finger $f$ from baseline prediction.

\subsection{3.2 Neighbour pixels voting}
We introduce our refinement stage in this section. After detecting stretching-out fingers, we aim at re-estimating locations of joints in detected stretching-out fingers, except the finger root which is in palm area. Briefly speaking, for each stretching-out fingers, the joints will get votes from foreground pixels around them. The average of votes each joint got is set to be its location after this refinement stage.

\subsubsection{3.2.1 Depth difference feature}\label{section_feature}
As proposed in \cite{shotton2013real}, depth difference features (Eq.\eqref{eq_feature}) are capable to distinguish background and foreground in depth images. Besides, different parts in hand will have different features. Also due to its mini cost in calculating, depth difference features are commonly used in hand pose estimation from depth images with random forests. We use such features in both cascaded hierarchical regression and neighbor pixels voting.
\begin{equation}\label{eq_feature}
  u_i = u + {\delta}u_i, i = 1, 2;\hspace{2mm}depth\hspace{1mm}difference\hspace{1mm}:= I(u_1) - I(u_2)
\end{equation}
Where u is the reference pixel, while specifically in our neighbor pixels voting, it is the foreground pixel which will vote for a joint's location. ${\delta}u_i, i = 1, 2$ are random offsets within a predefined range. $I(u_i), i = 1, 2$ are depths in pixel position $u_i$, which can be read from depth images directly.

\subsubsection{3.2.2 Training}
Our training algorithm is shown in Table.\ref{refine_train}.

\begin{table}[htb]
  \caption{Table.2. Training procedure of neighbor pixels voting}
  \label{refine_train}
  \begin{center}
  \begin{tabular}{p{0.02\columnwidth}p{0.96\columnwidth}}
    \hline
    \multicolumn{2}{l}{\emph{Algorithm 2}\hspace{2mm}Training algorithm for Neighbor Pixels Voting} \\
    \hline
    1. & \emph{input:} depth image $I_i$, and ground truth pose ${\theta}_i$ for randomly N samples in training set\\
    2. & \emph{training samples:} $\{I_i, pixel_{i, j}\}$, for $i = 1, 2, \cdots, N$, $pixel_{i, j}$ is a foreground pixel in $I_i$ \\
    3. & {\quad}$k^\ast = \arg\min_k\|pixel_{i, j} - joint_{i, k}\|_2$, \emph{find the nearest joint $(joint_{i, k^\ast})$ of $pixel_{i, j}$ in ${\theta}_i$} \\
    4. & {\quad}${\delta}\emph{x} = joint_{i, k^\ast} - pixel_{i, j}$ \\
    5. & {\quad}learn random forest $R$ to approximate ${\delta}\emph{x}$ \\
    6. & \emph{output:} $R$ \\
    \hline
  \end{tabular}
  \end{center}
\end{table}

\emph{Training samples collecting}{\qquad}Each foreground pixel $pixel_{i, j}$ in depth image $I_i$ is extracted as a training sample $\{I_i, pixel_{i, j}\}$. There are usually thousands of foreground pixels in one depth image, so we will extract thousands of training samples in each depth image. Considering the huge memory resume, we randomly select $N$ depth images in training set, ie. not all images are used in training procedure. $N$ has to be large enough, so selected depth images can roughly cover the distribution of hand poses in original training set.

All training samples collected in this way are fed into a random forest for training. We use depth difference feature (See Eq.\eqref{eq_feature} in Section.3.2.1) for split of tree nodes.

\emph{Prediction of leaf nodes}{\qquad}For each training sample $\{I_i, pixel_{i, j}\}$, we first find the nearest joint of $pixel_{i, j}$ in $\theta_i$ (Eq.\eqref{eq_nearest_joint}). The label of $\{I_i, pixel_{i, j}\}$ is the coordinate difference between $pixel_{i, j}$ and $joint_{i, k^\ast}$ (Eq.\eqref{eq_prediction}). Then the prediction of a leaf node is set to be the average of all labels of training samples reaching it.
\begin{equation}\label{eq_nearest_joint}
  k^\ast = \arg\min_k\|pixel_{i, j} - joint_{i, k}\|_2
\end{equation}
\begin{equation}\label{eq_prediction}
  {\delta}\emph{x} = joint_{i, k^\ast} - pixel_{i, j}
\end{equation}
Where $joint_{i, k}$ is the position of the $k^{th}$ joint in depth image $I_i$, and $joint_{i, k^\ast}$ is the position of the nearest joint of $pixel_{i, j}$ in depth image $I_i$.
\subsubsection{3.2.3 Testing}
After the detection of stretching-out fingers in Section.3.1, we refine predictions of only those joints in stretching-out fingers. Locations of other joints remain as predictions of our baseline in Section.2. Our testing procedure is shown in Table.\ref{refine_test}.

\begin{table}[htb]
  \caption{Table.3. Testing procedure of neighbor pixels voting}
  \label{refine_test}
  \begin{center}
  \begin{tabular}{p{0.02\columnwidth} p{0.96\columnwidth}}
    \hline
    \multicolumn{2}{l}{\emph{Algorithm 3}\hspace{2mm}Testing algorithm for Neighbor Pixels Voting} \\
    \hline
    1. & \emph{input:} depth image $I_i$ and hand pose $\theta^0_i$ after interpolation for all samples $i$ in testing set, random forest $R$ \\
    2. & {\quad}$dist = \min_k\|pixel_{i, j} - joint_{i, k}\|_2$, $pixel_{i, j}$ is a foreground pixel in $I_i$ \\
    3. & \emph{testing samples:} $\{I_i, pixel_{i, j}\}$, if $dist < thres_{dist}$, for all $i$, $thres_{dist}$ is a distance threshold \\
    4. & {\quad}$k^\ast = \arg\min_k\|pixel_{i, j} - joint_{i, k}\|_2$, \emph{find the nearest joint $(joint_{i, k^\ast})$ of $pixel_{i, j}$ in ${\theta}^0_i$}, for each testing sample \\
    5. & {\quad}$predict^{k^\ast, v}_{i, j} = R(I_i, pixel^v_{i, j})$, $pixel^v_{i, j}$ is the v-th foreground pixel voting for joint $k^\ast$ \\
    6. & for each updating joint $k^\ast$, update its location: \\
    \quad & ${\quad}joint^{new}_{i, k^\ast} = \frac{1}{V}\sum^V_{v=1}(pixel_{i, j} + predict^{k^\ast, v}_{i, j})$ \\
    7. & $\theta^{refine}_i = \{joint_{i, k}\}_{not\hspace{1mm}updated\hspace{1mm}joints\hspace{1mm}in\hspace{1mm}{\theta}^0_i} \bigcup \{joint_{i, k^\ast}\}_{updated\hspace{1mm}joints}$, \emph{update hand pose after refinement} \\
    8. & \emph{output:} $\theta^{refine}_i$ for all testing images \\
    \hline
  \end{tabular}
  \end{center}
\end{table}

The construction of testing sample set is similar as that in training. One difference is that not all foreground pixels are fed into random forest while testing. We first find the nearest joint of a certain foreground pixel $pixel_{i, j}$ in ${\theta}^0_i$, and calculate corresponding minimum distance. ${\theta}^0_i$ is the hand pose after interpolation in Section.3.1, locations of joints in stretching-out fingers are constrained within the detected finger contour. If the minimum distance is smaller than a predefined threshold, the foreground pixel is treated as a testing sample and fed into random forest for testing.

Each foreground pixel satisfied the mentioned distance threshold condition votes for the location of its nearest joint through random forest. In each tested depth image, a to-be-updated joint will obtain several votes from different pixels. We set the average of all votes got by a to-be-updated joint as its new location after refinement.

\section{4. Experiments And Discussions}\label{section_experiment}
In this section, we first introduce the evaluated datasets and metrics in our experiments. Then we compare results of our refinement and the baseline. Finally, we focus on the performance of proposed IHPE on interactive hand joint estimation, especially stretching-out finger tips.

\subsection{4.1 Experiment setup}
We show the parameters we used in neighbor pixels voting in Table.\ref{parameters}.

\begin{table}[htb]
  \caption{Table.4. Parameters we used in Neighbor Pixels Voting}
  \label{parameters}
  \begin{center}
  \begin{tabular}{|c|c|}
    \hline
    \emph{parameter} & \hspace{1mm}\emph{value}\hspace{1mm} \\\hline
    number of trees & 8 \\\hline
    maximum depth of trees & 20 \\\hline
    number of features for node split & 200 \\\hline
    \hspace{1mm}number of thresholds for feature division\hspace{1mm} & 50 \\\hline
    minimum information gain for node split & 1e-6 \\\hline
    minimum samples contained in a node & 5 \\\hline
    number of depth images for training & 10000 \\\hline
    distance threshold in testing (mm) & 10 \\
    \hline
  \end{tabular}
  \end{center}
\end{table}

\subsubsection{4.1.1 Datasets}
Just like \cite{sun2015cascaded}, we conducted our experiments on two publicly RGB-D datasets: ICVL hand pose dataset \cite{tang2014latent} and MSRA hand pose dataset \cite{sun2015cascaded}.

\emph{MSRA dataset}\hspace{4mm}The MSRA dataset contains 9 subjects with 17 gestures for each subject. 76.5K depth images with 21 annotated joints are collected with Intel's Creative Interactive Camera. Each subject is used for testing in turn while the remain eight subjects for training data. Totally 9 experiments are repeated, and the average result is reported.

\emph{ICVL dataset}\hspace{4mm}The training set of ICVL dataset contains 10 subjects with 26 gestures. About 22K depth images with 16 annotated joints are captured by Intel RealSense with the range of view about 120 degrees. The testing set contains 1.6K images.

\subsubsection{4.1.2 Evaluation metrics}
We employ several evaluation metrics following the literatures \cite{tang2014latent,tompson2014real}. For both MSRA dataset and ICVL dataset, the performance is evaluated by three metrics: (1) \emph{average 3D distance error} is calculated as the average Euclidean distance between ground truth and prediction for each joint (in millimeters). (2) \emph{all fingers error} is computed as the average Euclidean distance for joints in each finger (in millimeters). (3) \emph{all finger tips error} is defined as the average Euclidean distance for finger tip in each finger. For MSRA dataset, in which the depth images of 17 gestures are separated into different directories, we can evaluate the performance of our IHPE by two extra metrics: (4) \emph{stretching-out fingers error} is computed as the average Euclidean distance for joints in each finger (in millimeters), only those fingers stretching out are taken into account. (5) \emph{stretching-out finger tips error} is defined as the average Euclidean distance for finger tip in each finger, considering only those stretching-out fingers.

\subsection{4.2 Comparison with previous work}
We compare the experiment result after Stretching-out Finger Joint Refinement in Section.3 with the baseline in Section.2. Note that our baseline is achieved by re-implementing the cascaded framework and hierarchical regression in \cite{sun2015cascaded}, except that we use the traditional depth difference feature but not \emph{3d pose indexed feature} in \cite{sun2015cascaded}.

\begin{figure*}[htb]
\begin{minipage}[b]{0.49\textwidth}
  \centering
  \centerline{\includegraphics[width=0.95\textwidth]{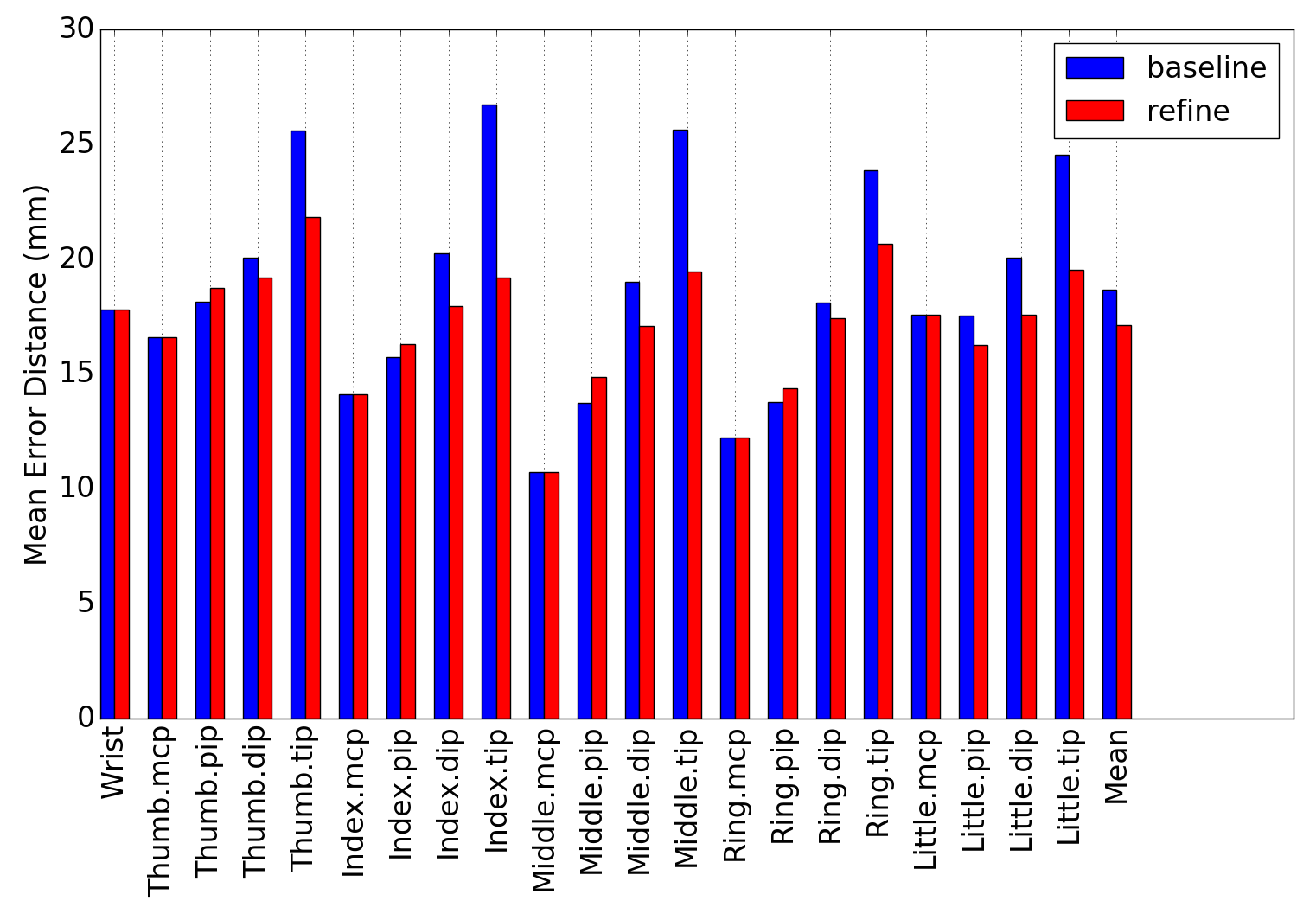}}
\end{minipage}
\begin{minipage}[b]{0.49\textwidth}
  \centering
  \centerline{\includegraphics[width=0.95\textwidth]{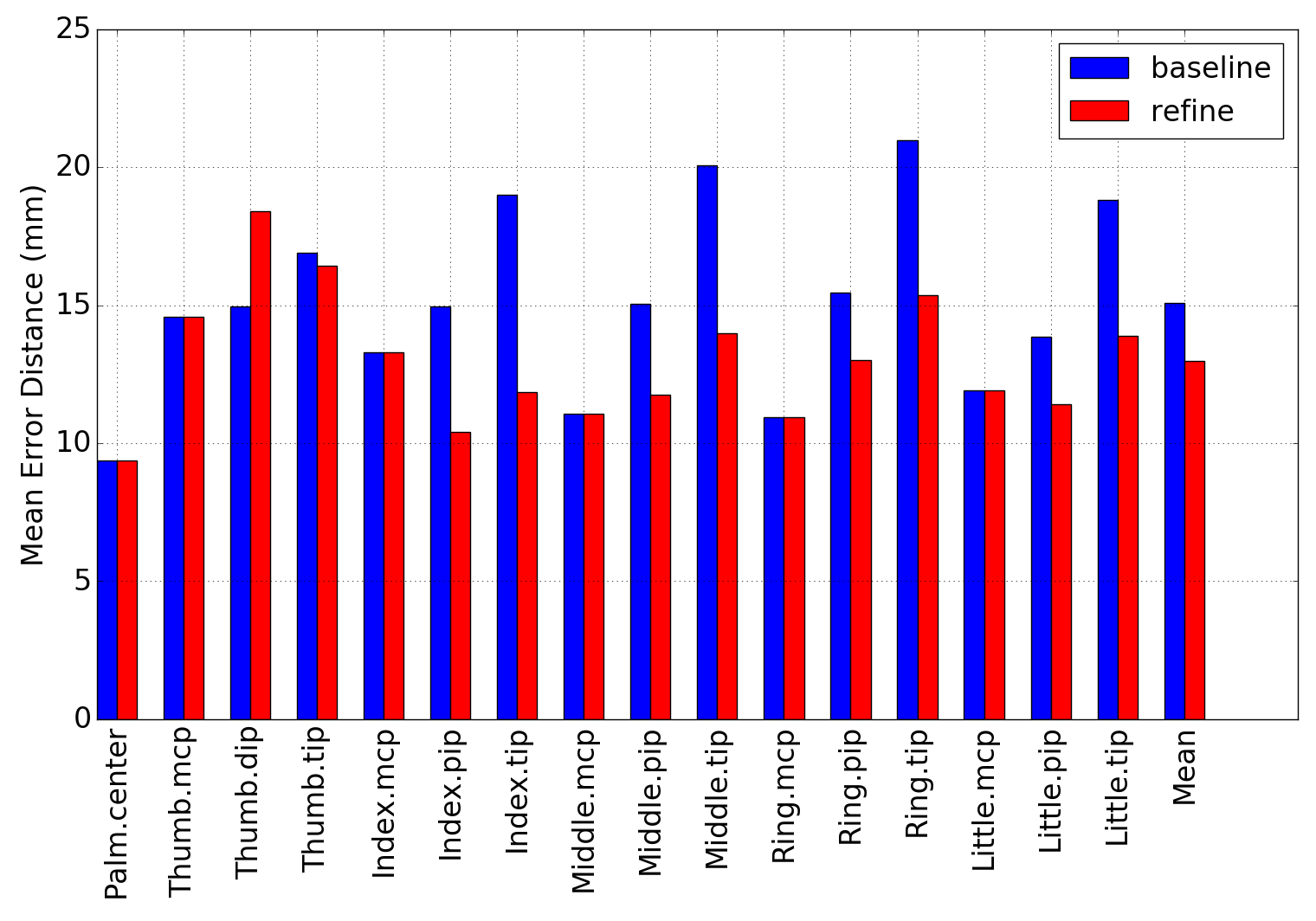}}
\end{minipage}
\caption{Distance error of our baseline\cite{sun2015cascaded} and refinement on MSRA\cite{sun2015cascaded} dataset (left), and ICVL\cite{tang2014latent} dataset (right)}
\label{fig_3derror}
\end{figure*}

The average 3D distance error on MSRA and ICVL datasets are shown in Fig.\ref{fig_3derror}. Mean error of all joints on MSRA is 18.65mm for baseline, and 17.11mm for refinement. Results on ICVL dataset are 15.07mm and 12.98mm respectively. The increasing error of the middle joint in thumb on ICVL is probably caused by the definition of this joint, which is almost dropped in the palm area. We can see significant improvements on the predictions of finger tip locations on both datasets.

Our refinement of stretching-out fingers achieves pleasing performance. This is beneficial from the the pulling force of stretching-out fingers detection and neighbor pixels voting, which drags those joints with large error in the baseline towards more precise areas inside fingers.

\subsection{4.3 Improvements in interactive joint estimation}
We show the mean of five fingers (or five finger tips) of \emph{all fingers error} and \emph{all finger tips error} of MSRA dataset and ICVL dataset in Table.\ref{table_msra_icvl}.

No matter \emph{all fingers error} or \emph{all finger tips error}, there is significant improvement on both datasets. We emphasize two conclusions here: (1)All finger tips error reduces 5.14mm on MSRA dataset (relatively 20.3\%) and 4.85mm on ICVL dataset (relatively 25.3\%). (2)The average error gap after refinement between fingers and fingertips is smaller than the baseline. This means that the conflict between HMI demand and model performance is alleviated. Locations of fingertips can be predicted as precise as other finger joints.

\begin{figure*}[htb]
\begin{minipage}[b]{0.49\textwidth}
  \centering
  \centerline{\includegraphics[width=0.95\textwidth]{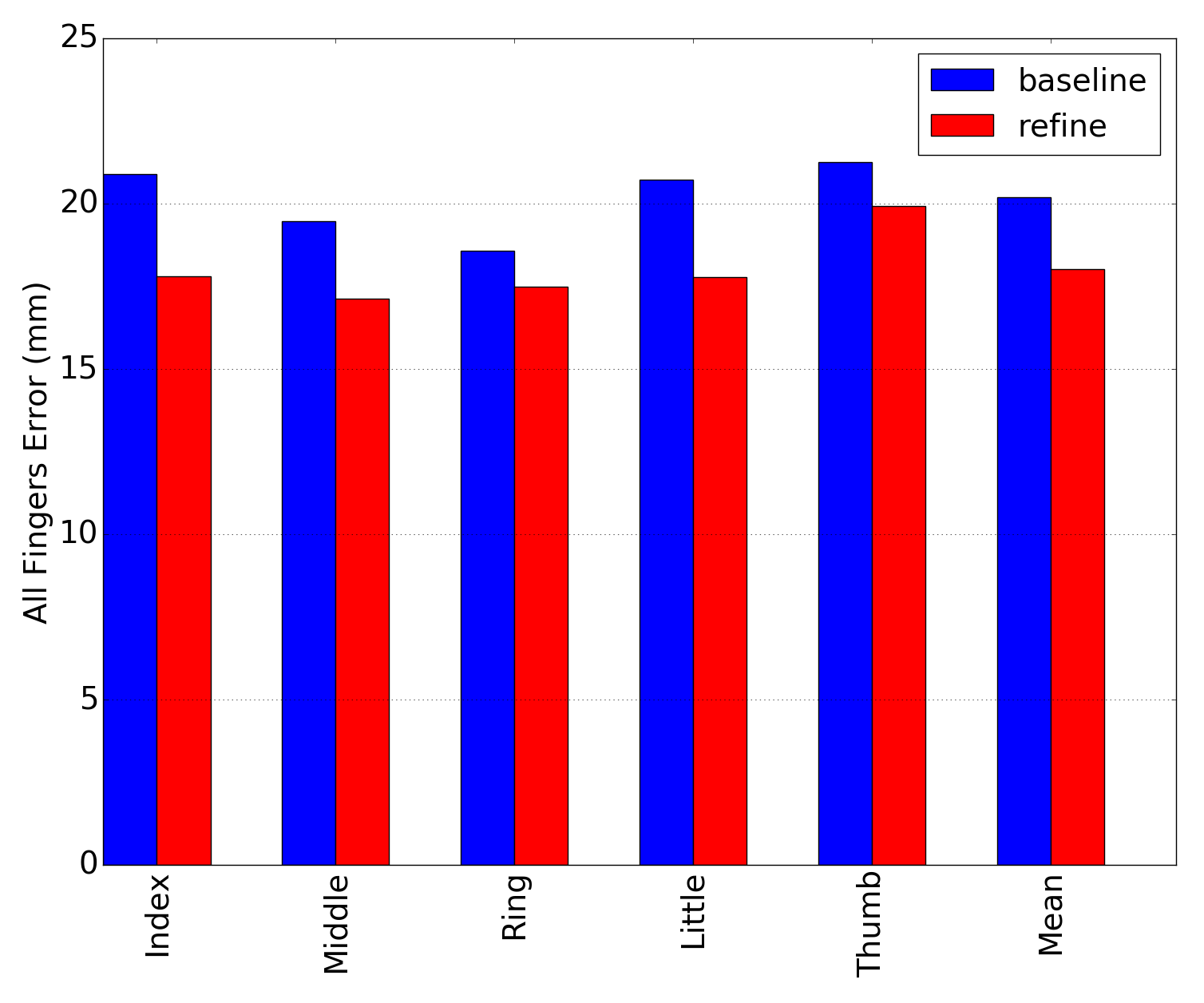}}
\end{minipage}
\begin{minipage}[b]{0.49\textwidth}
  \centering
  \centerline{\includegraphics[width=0.95\textwidth]{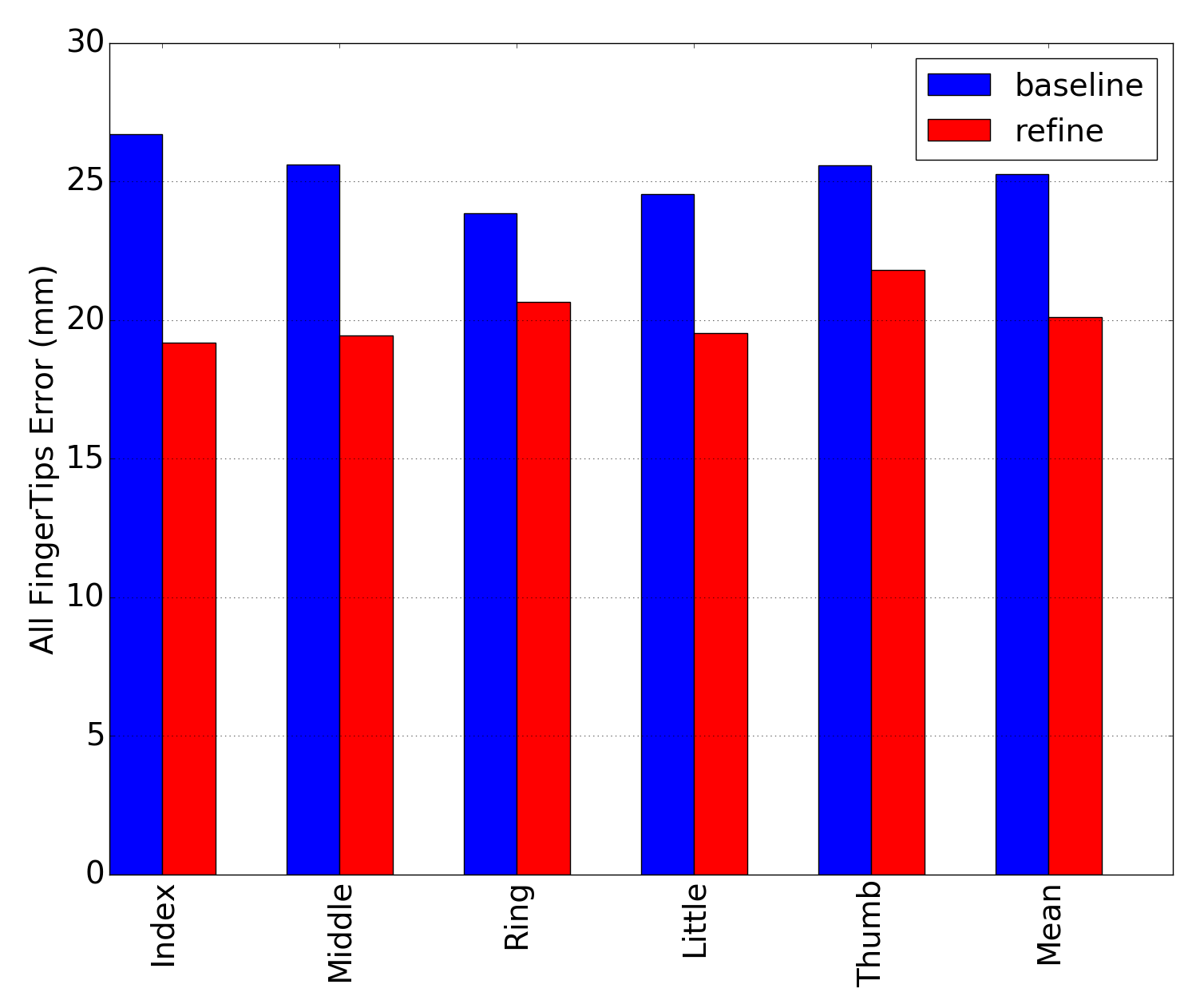}}
\end{minipage}
\caption{Improvements in finger joint estimation on MSRA\cite{sun2015cascaded} dataset: all fingers error (left) and all finger tips error (right).}
\label{fig_msra_all}
\end{figure*}

The results of \emph{all fingers error} and \emph{all finger tips error} of MSRA dataset are shown in Fig.\ref{fig_msra_all}.

\begin{table}[htb]
\caption{Table.5. Average of \emph{all fingers error} and \emph{all finger tips error}}
\label{table_msra_icvl}
\begin{center}
\begin{tabular}{|c|c|c|c|c|}
  \hline
  dataset & \multicolumn{2}{|c|}{MSRA} & \multicolumn{2}{|c|}{ICVL} \\\hline
  method & baseline & refine & baseline & refine \\\hline
  all fingers (mm) & 20.18 & 18.02 & 17.00 & 13.65 \\\hline
  all fingertips (mm) & 25.26 & 20.12 & 19.15 & 14.30 \\
  \hline
\end{tabular}
\end{center}
\end{table}

\begin{figure*}[htb]
\begin{minipage}[b]{0.49\textwidth}
  \centering
  \centerline{\includegraphics[width=0.95\textwidth]{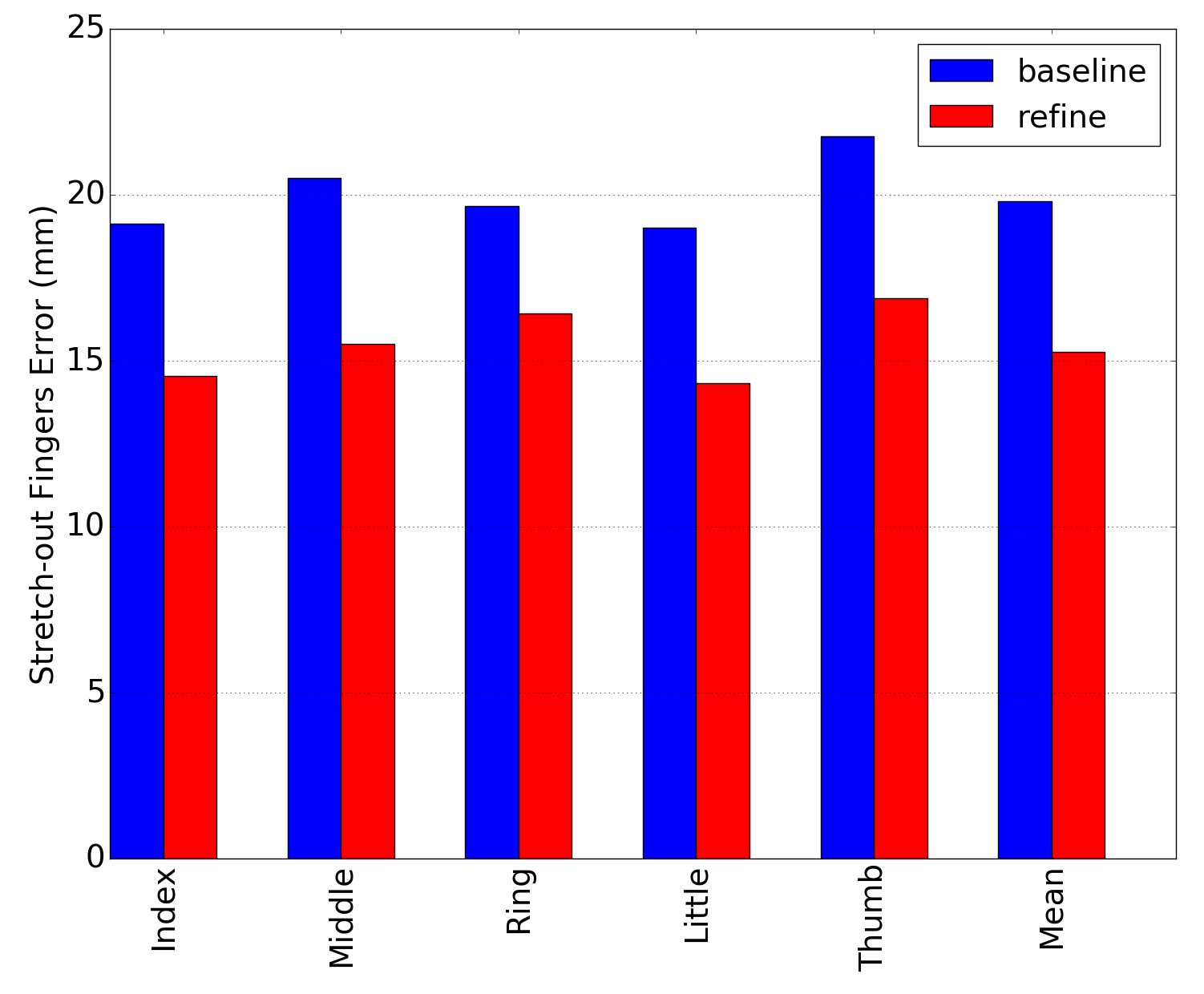}}
\end{minipage}
\begin{minipage}[b]{0.49\textwidth}
  \centering
  \centerline{\includegraphics[width=0.95\textwidth]{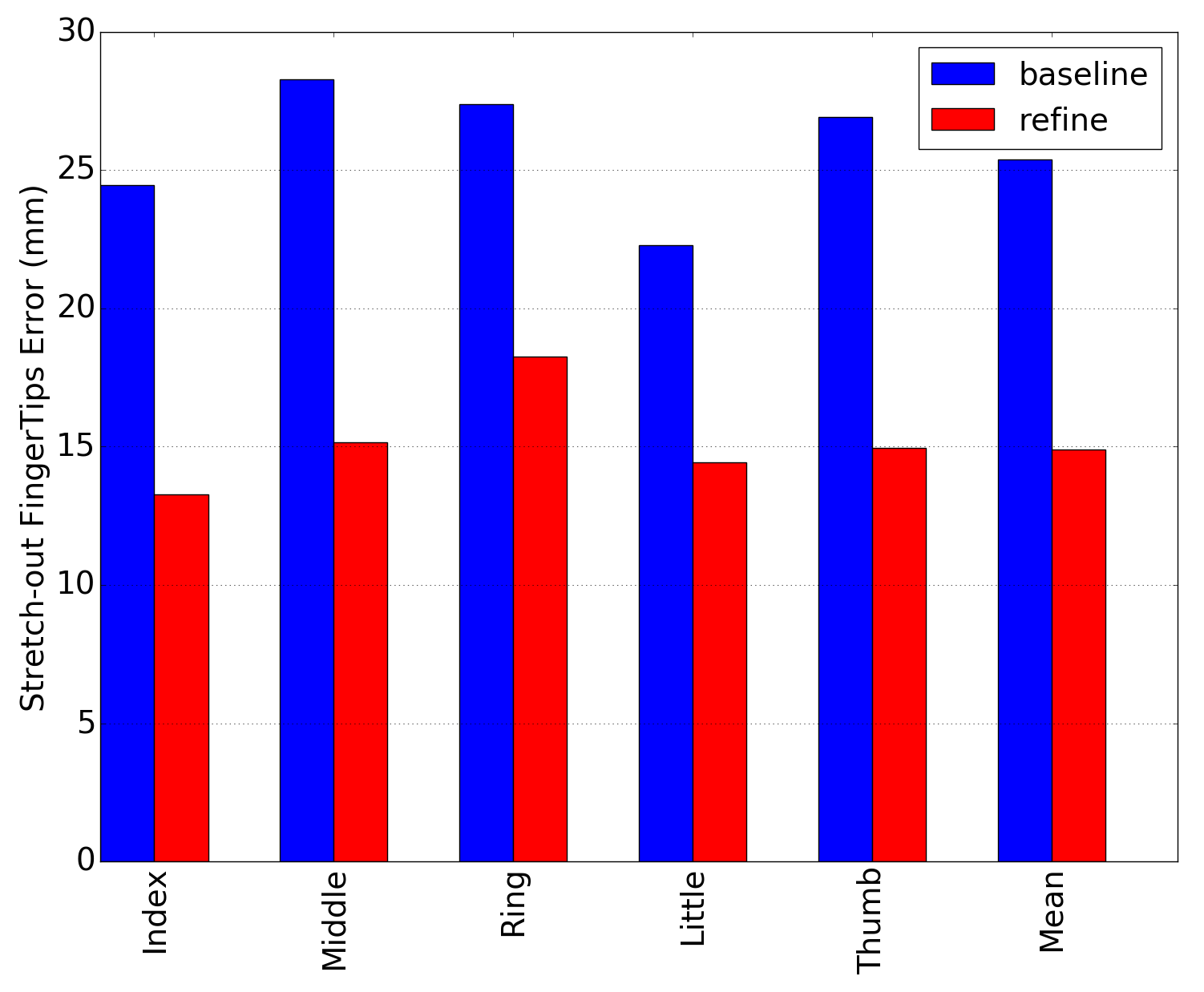}}
\end{minipage}
\caption{Improvements in interactive finger joint estimation on MSRA\cite{sun2015cascaded} dataset: stretching-out fingers error (left) and stretching-out finger tips error (right).}
\label{fig_msra_stretch}
\end{figure*}

Stretching-out fingers error and stretching-out fingertips error on MSRA are shown in Fig.\ref{fig_msra_stretch}. The average result of five fingers is listed in Table.\ref{table_msra_stretch}. Compared with the baseline, the average stretching-out fingers error reduces 4.54mm (relatively 22.9\%) after refinement. More significant improvement can be seen in the average stretching-out fingertips error, which reduces 10.51mm (relatively 41.4\%). In the reality of HMI, the most important joints are those tips of stretching-out fingers. Our IHPE system achieves more satisfying performance on hand pose estimation.

\begin{table}[htb]
\caption{Table.6.Average of \emph{stretching-out fingers error} and \emph{stretching-out finger tips error} on MSRA\cite{sun2015cascaded} dataset}
\label{table_msra_stretch}
\begin{center}
\begin{tabular}{|c|c|c|}
  \hline
  method & baseline & refine \\\hline
  stretching-out fingers (mm) & 19.81 & 15.27 \\\hline
  stretching-out fingertips (mm) & 25.39 & 14.88 \\
  \hline
\end{tabular}
\end{center}
\end{table}

\begin{figure}[htb]
  \begin{minipage}[htb]{0.49\columnwidth}
    \centering
    \includegraphics[width=0.95\columnwidth]{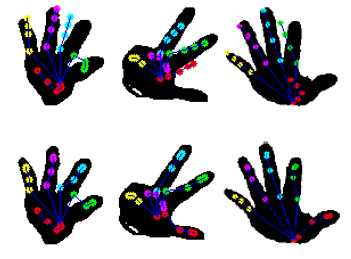}
    \caption{Example results of MSRA\cite{sun2015cascaded}: baseline (top), refinement (bottom). Note that after refinement, the wrist and finger roots remain the same as the baseline.}
    \label{fig_msra_examples}
  \end{minipage}
  \hspace{2mm}
  \begin{minipage}[htb]{0.49\columnwidth}
    \centering
    \includegraphics[width=0.95\columnwidth]{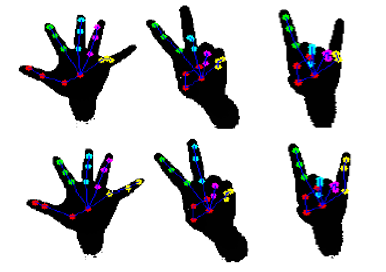}
    \caption{Example results of ICVL\cite{tang2014latent}: baseline (top), refinement (bottom). Note that after refinement, palm center and finger roots remain the same as the baseline.}
    \label{fig_icvl_examples}
  \end{minipage}
\end{figure}

Fig.\ref{fig_msra_examples} and Fig.\ref{fig_icvl_examples} show some examples of predictions of the baseline and corresponding one after refinement. The predictions of stretching-out finger joint locations are far more accurate after refinement.

\emph{Run Time}\hspace{4mm}We run the testing procedure of the whole IHPE system in a single thread at $Intel\circledR\hspace{1mm}Core^{TM}\hspace{1mm}i7-7700K\hspace{1mm}CP\hspace{1mm}@\hspace{1mm}4.20GHz\times8$. The overall speed achieves $18fps$, almost meeting real-time requirements.

\section{5. Conclusion}\label{section_conclusion}
We present a novel scheme named Interactive Hand Pose Estimation (IHPE) to address the conflicts between HMI demands and large error of fingertip locations in hand pose estimation. After obtaining rough locations of hand joints, we detect which fingers are stretching out based on the rough locations and depth images. For each joint in each stretching-out finger, its neighbor foreground pixels vote for its new location through random forests, and its final location is set to the average of voting results. As shown in our experiments, after the refinement stage of stretching-out finger joints, the errors of stretching-out finger joint locations are significantly reduced, especially the fingertips. The performance on ICVL dataset is good enough to perform rough Human Machine Interaction, although it needs to be better for more delicate operations, maybe smaller than 10mm for the error of localizing fingertips. Our future researches focus on further improving the accuracy, include raising the accuracy of detection of stretching-out fingers, and improving the performance of neighbor pixels voting by introducing weighted voting mechanism.

\section{Acknowledgments}
%add the acknowledgement section here
This work was partially supported by State High-Tech Research and Development Program of China (863 Program, No. 2015AA016304).


%%%%%%%%%%%%%%%%%%%%%%%%%%%%%%%%%%
% Bibliography
%%%%%%%%%%%%%%%%%%%%%%%%%%%%%%%%%%

\small
%\nocite{*}
\bibliographystyle{vipc}
\bibliography{refs}% Produces the bibliography via BibTeX.


\begin{biography}
\textbf{Cairong Zhang} received his B.S. degree from Department of Electronic Engineering, Tsinghua University, Beijing, China, in 2017, where he is currently working toward his M.S. degree. His research interests include human pose estimation and hand pose estimation.

\textbf{Guijin Wang} received his B.S. and Ph.D. degrees (with honor) from Tsinghua University, China, in 1998 and 2003 respectively, all in electronic engineering. From 2003 to 2006, he was a researcher at Sony Information Technologies Laboratories. Since October 2006, he has been with the Department of Electronic Engineering, Tsinghua University, China, as an associate professor. His research interests focus on wireless multimedia, depth sensing, pose recognition, intelligent human-machine UI, intelligent surveillance, industry inspection, and online learning.

\textbf{Hengkai Guo} received his B.S. and M.S. degree from Tsinghua University, Beijing, China, in 2014 and 2017 respectively. His research interests include human pose estimation, hand pose estimation and deep learning.

\textbf{Xinghao Chen} received his B.S. degree from Department of Electronic Engineering, Tsinghua University, Beijing, China, in 2013, where he is currently pursuing the Ph.D. degree. From Sept. 2016 - Jan. 2017, he was a visiting Ph.D. student with Imperial College London, UK. His research interests include deep learning, hand pose estimation and gesture recognition.

\textbf{Fei Qiao} received his B.S. degree from Lanzhou University, China, in 2000, and his Ph.D. degree from Tsinghua University, Beijing, China, in 2006, both in electronic engineering. He is now an associate professor in Department of Electronic Engineering in Tsinghua University. His research interests include Smart Camera Architectures for Depth Perception Applications, X Computing Architectures and Heterogeneous Integration of Electronic-Eyes.

\textbf{Huazhong Yang} received his B.S. and Ph.D. degree from Department of Electronic Engineering, Tsinghua University, Beijing, China, in 1993 and 1998 respectively. Since July 1998, he has been with the Department of Electronic Engineering, Tsinghua University, China, as a professor. His research interests focus on new structure, synthesis and verification of micro-system chip, and design of analog and mixed-signal system.

\end{biography}

\end{document}